\documentclass{bmvc2k}
 
\usepackage{amsmath,amssymb}
\usepackage{graphicx}
\usepackage{booktabs}
\usepackage{microtype}
\usepackage{float}
\usepackage{hyperref}

\title{ActiveAugment: Online Active Learning for Augmentation Selection in Deep Learning}
\addauthor{Noah Videcrantz}{novi@di.ku.dk}{1}
\addauthor{Mostafa Mehdipour Ghazi}{ghazi@di.ku.dk}{1}
\addinstitution{Pioneer Centre for AI\\University of Copenhagen}
\runninghead{Videcrantz and Mehdipour Ghazi}{Active--Augment}

\def\eg{\emph{e.g}\bmvaOneDot}
\def\ie{\emph{i.e}\bmvaOneDot}

\begin{document}

\maketitle

\begin{abstract}
Data augmentation is a cornerstone of deep learning pipelines, yet existing strategies treat it as a static, model-agnostic preprocessing step, either relying on expensive dataset-specific policy search or applying transformations uniformly at random, regardless of what the model has already learned. We introduce ActiveAugment, a unified framework that treats augmentation selection as an online active learning problem. For each training minibatch, ActiveAugment generates a pool of candidate augmented views and scores each candidate using a combination of the model's predictive uncertainty and the feature discrepancy induced by the augmentation. The augmentation under which the current model is most fragile is selected per sample, and the model is then trained with a joint supervised classification and supervised contrastive objective that enforces intra-class invariance to the selected augmentations while maintaining inter-class separation. We evaluate ActiveAugment on eight benchmark datasets spanning natural and medical imaging, using CNN and transformer architectures across three training regimes (training from scratch, full fine-tuning, and linear probing), and comparing eight active selection strategies for augmentation scoring. ActiveAugment outperforms AutoAugment, RandAugment, and TrivialAugment under controlled augmentation shifts across all domains and budgets, with the most pronounced gains at low labelling budgets. On medical imaging datasets, where data is scarce and domain shift relative to natural-image pretrained models is large, ActiveAugment achieves higher test F1 than all baselines, demonstrating strong cross-domain adaptability. Our analysis reveals that the augmentation selection policy evolves meaningfully during training and that strategy choice has a direct impact on generalisation. Code is available at: \url{https://github.com/noahvide/ActiveAugment}.
\end{abstract}

\section{Introduction} \label{sec:intro}

Training deep learning models typically requires large volumes of labelled data, and this need is only intensifying as models grow in size and complexity \cite{wang2024computation}. Yet, acquiring large amounts of annotated data is not always feasible due to constraints such as costly expert labelling or access to biased corpora \cite{chi2020deep,vadineanu2024}. Consequently, data-efficient deep learning has become an essential research direction aimed at maximising the utility of limited available samples \cite{wang2021selftuning}.

Two complementary strategies have emerged for maximising data efficiency. \emph{Active Learning} (AL) reduces the number of required labels by identifying the most informative instances from an unlabelled pool for annotation \cite{settles2009}. \emph{Data augmentation} increases the effective utility of existing labels by synthesising additional training views \cite{shorten2019survey}. While data augmentation is a cornerstone of modern training pipelines, it is treated as a static step rather than an adaptive component of the learning process, a mismatch that motivates this work.

Several challenges arise when designing an augmentation pipeline. A central one is choosing \emph{which} augmentations to apply: a horizontal flip may be label-preserving for many natural-image tasks yet label-destructive in others (\eg digit recognition or anatomically oriented medical imaging tasks). Another challenge is deciding on the \emph{intensity} of each transformation. Furthermore, as the model learns, it may become invariant to some augmentations while remaining fragile to others, so a static pipeline becomes increasingly misaligned with the model's current state. Given the combinatorial space of augmentation types and intensities, together with the dynamic nature of training, there is a need for improved augmentation.

Automatic augmentation selection has been explored previously. AutoAugment \cite{cubuk2019autoaugment} treats the problem as a reinforcement-learning search over a discrete policy space and yields a strong, dataset-specific policy. However, the search is computationally expensive and the resulting policy is static, making it model-agnostic. RandAugment \cite{cubuk2020randaugment} reduces the search space to two scalar hyperparameters (pipeline length and global intensity), making the search cheaper while achieving comparable accuracy. Nevertheless it remains dataset-specific and model-agnostic. TrivialAugment \cite{muller2021trivialaugment} eliminates the search entirely by sampling a single augmentation and intensity uniformly at random for each image, achieving state-of-the-art results on several benchmarks. While dataset-agnostic, it is also model-agnostic. This leaves a gap between simple, model-blind random sampling and expensive, static policy searches.

The technical core of adaptive augmentation selection lies in identifying the model's empirical blind spots. Rather than treating data augmentation purely as a heuristic tool for dataset expansion, it can be formally viewed through the lens of domain adaptation: clean training data acts as the source distribution, and augmented variants represent a proxy target domain. Classic generalisation bounds establish that target-domain performance is upper-bounded by the source error and the divergence between the two feature distributions \cite{ben2010theory,mansour2009domain}. Uniform or random augmentation strategies fail to optimise this bound because they apply transformations blindly. Without a mechanism to identify where the model is fragile, training steps are spent on uninformative augmentations that do not tighten the generalisation gap.

We propose that augmentation selection should be treated as an \emph{active} process, using informativeness measures from the AL literature to select the most beneficial augmentations online during training. Therefore, we introduce \textbf{ActiveAugment}, a framework for active invariance learning via selective augmentation. The framework decouples training into two phases: a \emph{Selection} phase and a \emph{Learning} phase. During selection, a pool of candidate augmented views is generated; each candidate is scored by a combination of the model's predictive uncertainty and the feature discrepancy induced by the augmentation. This allows the model to identify and prioritise augmentations under which it is currently most fragile. During learning, the selected augmentations are used within a supervised contrastive objective that simultaneously enforces intra-class invariance and inter-class separation.

We evaluate ActiveAugment using convolutional and transformer backbones across three training regimes (training from scratch, full fine-tuning, and linear probing) on eight datasets, including natural image benchmarks (CIFAR-10/100, STL-10, MNIST) and medical imaging datasets (BRISC, BUSI, FETAL-PLANES-DB, ISIC-2019). Our main contributions are:
\begin{itemize}
\item A novel unified framework, ActiveAugment, that reformulates augmentation selection as an online AL problem, combining predictive uncertainty and feature discrepancy into a principled, theoretically grounded selection score.
\item A joint training objective that pairs active augmentation selection with supervised contrastive learning, enforcing invariance to model-fragile augmentations while maintaining inter-class discriminability.
\item A systematic translation of eight AL strategies (uncertainty-based, diversity-based, and hybrid) into online augmentation-scoring strategies, with detailed analysis of their selection behaviour and its performance consequences.
\item Comprehensive empirical evaluation across eight datasets, two backbone architectures, and three training regimes, demonstrating consistent gains over AutoAugment, RandAugment, and TrivialAugment, particularly at low labelling budgets and in the medical imaging domain where domain-adaptive augmentation selection is most beneficial.
\end{itemize}

\section{Related Work} \label{sec:related}

\subsection{Data Augmentation} \label{sec:related_aug}

Data augmentation addresses two central challenges in supervised learning: insufficient training data and overfitting. By augmenting samples and appending them to the training pool, one artificially increases the dataset size; by applying a different transformation each time a sample is presented, the network is forced to rely on robust, semantically meaningful features rather than surface-level memorisation.

The earliest augmentations were geometric transformations applied when training LeNet-5 \cite{lecun1998gradient}, including flipping, shearing, and rotation. When training AlexNet, photometric augmentations altering pixel-level properties like brightness and contrast were introduced alongside geometric ones \cite{krizhevsky2012imagenet}. Subsequent work added augmentations like Cutout \cite{devries2017cutout} and Mixup \cite{zhang2018mixup}, further expanding the augmentation repertoire. These augmentations were selected largely by heuristics, motivating the development of automatic augmentation methods.

AutoAugment \cite{cubuk2019autoaugment} was the first to frame augmentation selection as a search problem. It parameterises the policy search space with a recurrent neural network controller trained via policy gradient on a small proxy task. While effective, the search is computationally expensive.  RandAugment \cite{cubuk2020randaugment} reduced the search space to two hyperparameters, achieving comparable results at a fraction of the cost; augmentations are drawn uniformly at random from a fixed pool, all at the chosen intensity. TrivialAugment \cite{muller2021trivialaugment} removed the need for search entirely: for each sample, it uniformly samples a single augmentation type and a random intensity, achieving strong results with zero hyperparameter tuning.

Related approaches explicitly optimize augmentation hardness. MaxUp \cite{gong2021maxup} generates multiple randomly augmented views and trains on the view with the largest loss, whereas AugMax \cite{wang2021augmax} constructs adversarial mixtures of randomly sampled augmentations to improve robustness. ActiveAugment differs by framing augmentation choice as online active selection: candidate views are scored using predictive informativeness and their relative representation shift, and standard AL acquisition strategies translate to the augmentation-selection problem.

\subsection{Active Learning} \label{sec:related_al}

Active learning is a framework designed to maximise data efficiency under labelling-budget constraints by querying the most informative instances from an unlabelled pool \cite{settles2009}. Two key considerations govern the selection of samples to label. First, one should prioritise \emph{uncertain} samples, those for which the current model is least confident, since labelling them yields the largest expected reduction in generalisation error.  Methods operating on this principle are called uncertainty-based methods. Second, selecting a batch of highly similar uncertain samples is wasteful; the selection should also be \emph{diverse} so that it covers the available input space. Methods operating on this principle are called diversity-based methods.

Among uncertainty-based methods, Least Confident sampling \cite{settles2009} selects the instances whose maximum predicted class probability is lowest. Margin Sampling focuses on the decision boundary, prioritising instances whose two highest class probabilities are closest. Bayesian Active Learning by Disagreement (BALD) \cite{gal2017deep} uses Monte Carlo Dropout to estimate the mutual information between predictions and model parameters. Coreset \cite{sener2018active} and TypiClust \cite{hacohen2022active} are representative diversity-based methods: Coreset selects samples that maximise the minimum distance to already-labelled instances, ensuring high coverage, while TypiClust uses clustering to find typical representatives that avoid overly difficult samples in early rounds. Hybrid approaches combine both criteria; BADGE \cite{ash2020deep} operates in gradient-embedding space, using gradient magnitude as a proxy for uncertainty and gradient direction as a proxy for diversity, selecting a batch that is simultaneously uncertain and diverse.

Although AL methods were originally formulated for unlabelled sample selection, most methods do not inherently depend on whether the informativeness measure is applied to a novel image or an augmented view of an existing one. This opens the door to translating sample selection directly into augmentation selection, which is the central idea of this work.

\vspace{-0.1cm}
\subsection{Contrastive and Invariance Learning} \label{sec:related_cl}

Actively selecting informative augmentations provides the model with challenging training views; however, learning robust representations from those views requires an appropriate objective. Standard cross-entropy loss is susceptible to shortcut learning, where the model exploits superficial correlations that do not generalise \cite{geirhos2020shortcut}. Contrastive learning methods address this by restructuring the geometry of the latent space: representations of related samples are pulled together while those of unrelated samples are pushed apart. SimCLR \cite{chen2020simple} and MoCo \cite{he2020momentum} are influential \emph{self-supervised} contrastive frameworks. Each sample is augmented in two independent ways to produce a positive pair; the model is optimised to maximise agreement within positive pairs while repelling other samples in the batch as negatives. While effective for unsupervised training, instance-discrimination objectives treat instances of the same class as negatives, which is suboptimal in supervised learning.

Supervised Contrastive Learning (SupCon) \cite{khosla2020supervised} addresses this limitation: it treats all augmented views that share the same class label as positives, while views from different classes serve as negatives. This enforces intra-class invariance while maintaining inter-class separation. Existing implementations sample positive augmented views at random, which may spend computation on views that provide little additional learning signal. ActiveAugment instead prioritises informative views from a predefined, domain-valid augmentation space.

\section{Method} \label{sec:method}

\subsection{Augmentation Space} \label{sec:augmentations}

We define an augmentation as a parameterised function $a^{I}(x)$, where $x$ is an input image and $I \in \{\texttt{L},\, \texttt{M},\, \texttt{H}\}$ denotes the intensity level (low, medium, high). Since each intensity level corresponds to a \emph{range} of raw transformation strengths, the exact magnitude is sampled uniformly at random from within that range at each application. Most augmentations are controlled by a single continuous strength parameter; augmentations such as random horizontal flipping have no associated magnitude and therefore no intensity level. The intensity ranges for each augmentation are fixed before training. We use heuristics to choose initial ranges and verify them using the Structural Similarity Index Measure (SSIM) \cite{wang2004ssim} to confirm that each level produces a qualitatively distinct degree of image corruption. Note that intensity does not map monotonically to perceptual corruption for all augmentations: for example, low Brightness yields a dark image, while medium Brightness produces an image close to the original.

% A complete list of augmentations and their intensity ranges is provided in Supplementary Material. 
Note that the candidate space is defined before training using domain knowledge; ActiveAugment therefore assumes, rather than automatically verifies, that candidate transformations are label-preserving for the task. SSIM is used only to verify distinct perturbation levels. We sample the low/medium/high intensity during selection for tractability, although the framework can naturally extend to joint augmentation–intensity selection. This discrete intensity parameterisation limits the effective search space to $|\mathcal{A}| \times 3$ (minus the intensity-free augmentations), making online selection tractable.

\subsection{Active Invariance Learning via Selective Augmentation} \label{sec:method_main}

The goal is to select, at each training step, the augmentation under which the model is most fragile, and to enforce invariance to that augmentation during training. Let $\mathcal{D} = \{(x_i,y_i)\}_{i=1}^{N}$ be a labelled dataset, $f_\theta : \mathcal{X} \to \mathbb{R}^d$ an encoder parameterised by $\theta$, $g_\phi$ a classification head parameterised by $\phi$, and $\mathcal{A}$ the augmentation space. Classical invariance learning enforces:
\begin{equation}
    f_\theta(x) \approx f_\theta\!\left(a(x)\right), \quad \forall\, a \in \mathcal{A},\;
    \forall\, x \in \mathcal{X}.
\end{equation}
Enforcing invariance uniformly across $\mathcal{A}$ is suboptimal: augmentations affect the model disproportionately. Within a predefined domain-valid augmentation space, different transformations affect the model disproportionately: some expose model fragilities whereas others provide little additional learning signal. Hence, we decouple training into two phases:
\begin{itemize}
  \item \textbf{Selection:} For each sample $x_i$, identify the augmentation $a \in \mathcal{A}$ from the predefined domain-valid candidate space, under which the model is currently most non-invariant.
  \item \textbf{Learning:} Enforce invariance specifically to the selected augmentations.
\end{itemize}
We combine two complementary informativeness signals during selection.

\paragraph{Predictive uncertainty ($u_{ik}$).}

We define predictive uncertainty as a score indicating how far the model is from a confident, stable prediction when presented with augmented view $\hat{x}_{i,k} = a_k^I(x_i)$. Any AL acquisition function can serve this role; we adapt existing AL strategies to score augmented views rather than unlabelled samples (see Section \ref{sec:al_strategies}).

\paragraph{Feature discrepancy ($d_{ik}$).}

We define feature discrepancy as the distance between the representation of the original sample and that of its augmented counterpart in the encoder's embedding space:
\begin{equation}
    d_{ik} = \mathrm{D}\!\left(f_\theta(x_i),\, f_\theta(\hat{x}_{i,k})\right),
    \label{eq:discrepancy}
\end{equation}
where $\mathrm{D}$ is a distance metric such as cosine distance. This measures how much augmentation $a_k$ displaces the representation relative to the clean sample. To prevent the score from being dominated by globally large or small embedding norms, we normalise $d_{ik}$ by the mean discrepancy across samples for the same augmentation:
\begin{equation}
    \bar{d}_k = \frac{1}{B} \sum_{i=1}^{B} d_{ik},
\end{equation}
where $B$ is the minibatch size. The combined selection score is:
\begin{equation}
    s_{ik} = \frac{d_{ik}}{\bar{d}_k} \cdot u_{ik}.
    \label{eq:score}
\end{equation}
The selected augmentation for sample $x_i$ is:
\begin{equation}
    a_i^* = \operatorname*{arg\,max}_{k \in \{1,\ldots,K\}} s_{ik}.
    \label{eq:argmax}
\end{equation}

\paragraph{Theoretical motivation.}

The selection criterion in Eq. \eqref{eq:score} can be grounded in a generalisation bound under augmentation-induced distribution shift. Let $P_{\mathcal{X}}$ denote the clean data distribution and $P_{\mathcal{A}}$ the distribution over augmented inputs. The generalisation error of $f_\theta$ under augmentation shift is bounded by \cite{ben2010theory,mansour2009domain}:
\begin{equation}
    \varepsilon_{\mathcal{A}}(f_\theta) \leq \varepsilon_{\mathcal{X}}(f_\theta)
    + \underbrace{d_{\mathcal{H}}(P_{\mathcal{X}}, P_{\mathcal{A}})}_{\text{representation shift}}
    + \underbrace{\lambda^*}_{\text{ideal joint risk}},
    \label{eq:bound}
\end{equation}
where $d_{\mathcal{H}}$ is the $\mathcal{H}$-divergence between the feature distributions of clean and augmented data. The discrepancy term $d_{ik}$ is a sample-level estimator of this divergence; minimising $\mathbb{E}_i[d_{ik^*}]$ via the contrastive objective tightens the bound for the selected augmentations. By a standard AL argument \cite{settles2009}, training on samples where the model exhibits high predictive entropy yields the largest expected reduction in loss per update, as gradient contributions scale with prediction error. Combining both signals, the selection score $s_{ik}$ approximately maximises:
\begin{equation}
    \Delta\mathcal{L}_{ik} \;\propto\; \underbrace{u_{ik}}_{\text{expected loss reduction}}
    \cdot \underbrace{d_{ik}/\bar{d}_k}_{\text{relative shift contribution}},
\end{equation}
which can be interpreted as a per-sample, per-augmentation expected improvement in the generalisation bound. The greedy selection rule $a_i^* = \operatorname*{arg\,max}_k s_{ik}$ is a principled approximation to the problem of jointly optimising augmentation selection to maximise bound tightening.

\paragraph{Algorithm.}

\begin{figure}[t]
  \centering
  \includegraphics[width=0.86\linewidth,height=0.26\textheight]{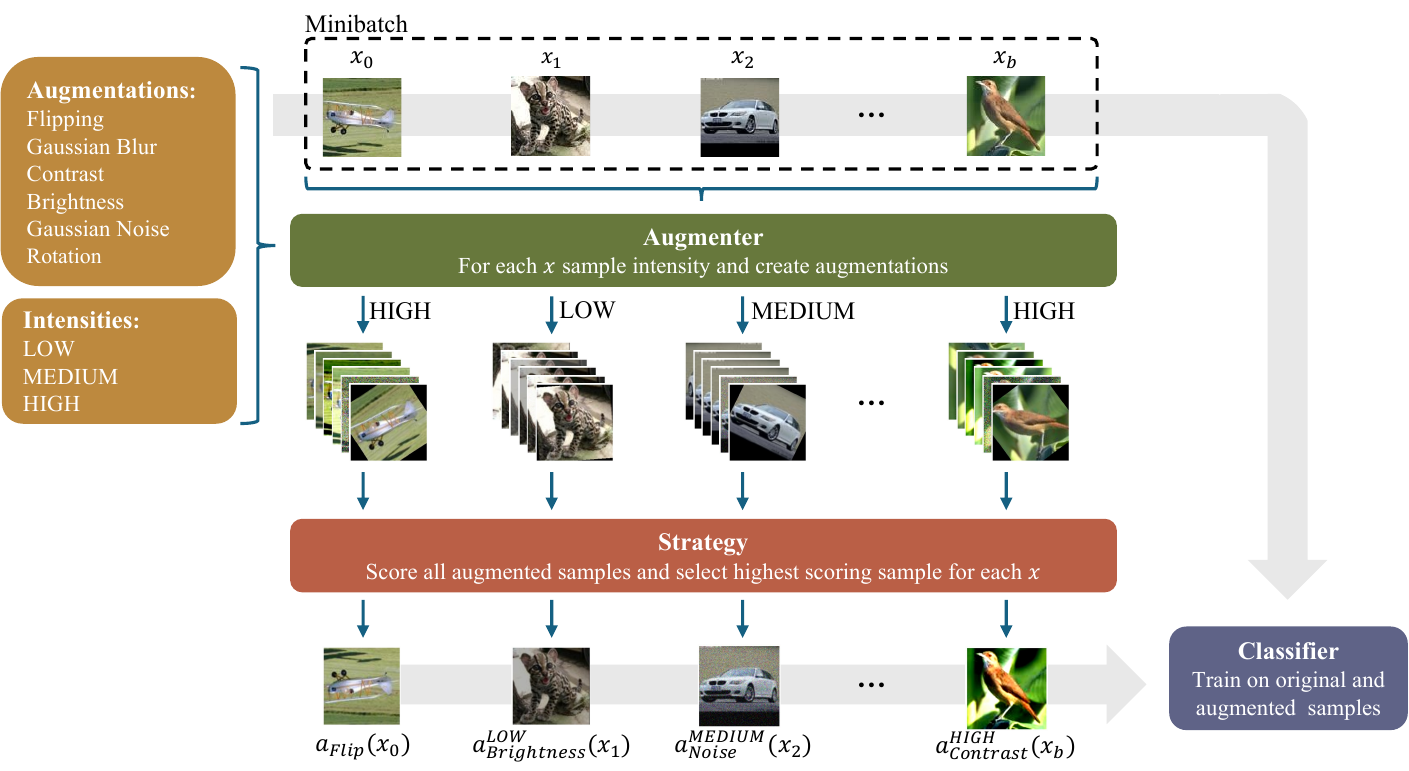}
  \caption{Overview of ActiveAugment. For each minibatch, every sample is augmented with all available augmentations at a randomly sampled intensity level. An active selection strategy scores each candidate and selects the most informative augmentation per sample. The model is then trained on both the original and selected augmented samples using the combined supervised classification and contrastive loss.}
  \label{fig:overview}
\end{figure}

Figure~\ref{fig:overview} provides an overview of ActiveAugment. For each minibatch $\mathcal{B} \subset \mathcal{D}$ of size $B$, we perform online selection as follows. For each sample $x_i \in \mathcal{B}$, we sample an intensity level $I \sim \mathcal{U}\{\texttt{L}, \texttt{M}, \texttt{H}\}$ and generate $K = |\mathcal{A}|$ candidate augmented views:
\begin{equation}
    \{\hat{x}_{i,k}\}_{k=1}^{K} = \left\{a_k^{I}(x_i) \;\middle|\; a_k \in \mathcal{A}\right\}.
\end{equation}
For each candidate $\hat{x}_{i,k}$ we compute the score $s_{ik}$ using the current model state, and select the highest-scoring augmentation:
\begin{equation}
    a_i^* = \operatorname*{arg\,max}_{k} \, s_{ik}.
\end{equation}
The model is then trained on the expanded minibatch $\mathcal{B}_{\mathrm{final}} = \{(x_i, y_i)\} \cup \{(a_i^*(x_i), y_i)\}_{i=1}^{B}$, updating encoder parameters $\theta$ and classifier parameters $\phi$.

\subsection{Selection Strategies} \label{sec:al_strategies}

AL provides the scoring framework for the selection phase. For \emph{uncertainty-based} strategies, the translation from sample scoring to augmentation scoring is straightforward: unlabelled samples are simply replaced by augmented views of labelled samples. This holds for Entropy Sampling, Least Confident Sampling, Margin Sampling, and BALD \cite{gal2017deep,settles2009}.

\emph{Diversity-based} strategies typically operate on the global labelled pool. We use the local clean minibatch as the reference distribution, making diversity computation tractable in the online setting. We apply this local adaptation to Coreset, BADGE, and TypiClust \cite{ash2020deep,sener2018active,hacohen2022active}.

For BAL \cite{li2024bal}, instead of performing global $k$-means clustering and scoring based on the Cluster Distance Difference, we compute a batch-adapted Cluster Distance Difference: for each augmented candidate, we measure its distance to its own clean source image versus its distance to the nearest other clean image in the minibatch. In total we implement and compare eight active learning strategies for augmentation scoring.

\subsection{Training Objective} \label{sec:loss}

We train the model with a combined supervised classification and supervised contrastive objective, with the balance controlled by hyperparameter $\lambda > 0$:
\begin{equation}
    \mathcal{L} = \mathcal{L}_{\mathrm{sup}} + \lambda \, \mathcal{L}_{\mathrm{con}}.
    \label{eq:loss}
\end{equation}
This joint objective prevents feature collapse, encourages invariance to the selected augmentations within classes, and maintains discriminative structure across classes.

\paragraph{Supervised classification loss.}

$\mathcal{L}_{\mathrm{sup}}$ is the standard cross-entropy loss $\ell$ applied to both the original samples and their selected augmented counterparts:
\begin{equation}
    \mathcal{L}_{\mathrm{sup}} = \frac{1}{B} \sum_{i=1}^{B}
        \Bigl[\ell\!\left(g_\phi(f_\theta(x_i)),\, y_i\right)
        + \ell\!\left(g_\phi(f_\theta(a_i^*(x_i))),\, y_i\right)\Bigr].
\end{equation}

\paragraph{Supervised contrastive loss.}

$\mathcal{L}_{\mathrm{con}}$ is the SupCon loss \cite{khosla2020supervised} applied over the expanded minibatch of size $2B$ (original plus augmented samples), using cosine similarity as the kernel:
\begin{equation}
    \mathcal{L}_{\mathrm{con}} = -\frac{1}{2B} \sum_{i=1}^{2B}
    \frac{1}{|P(i)|} \sum_{j \in P(i)}
    \log \frac{\exp\!\left(\mathrm{sim}(z_i, z_j)/\tau\right)}
              {\sum_{n=1,\, n \neq i}^{2B} \exp\!\left(\mathrm{sim}(z_i, z_n)/\tau\right)},
    \label{eq:supcon}
\end{equation}
where $z_i = g_\phi(f_\theta(x_i))$, $\mathrm{sim}(u, v) = u^\top v / (\|u\|\|v\|)$ is cosine similarity, $\tau > 0$ is a temperature hyperparameter, and $P(i) = \{j \neq i : y_j = y_i\}$ is the set of positive pairs, \ie all samples in the expanded minibatch that share the same class label as sample $i$. The SupCon loss enforces intra-class invariance while maintaining inter-class separation. In this work, $g_\phi$ serves as the classification head for $\mathcal{L}_{\mathrm{sup}}$ and the projection for $\mathcal{L}_{\mathrm{con}}$; a unified head simplifies the architecture and was found empirically sufficient for the scales studied here.

\subsection{Backbone Architectures} \label{sec:backbones}

We evaluate ActiveAugment using two backbone architectures that represent the two dominant paradigms in modern image recognition: \textbf{ResNet-18} \cite{he2016deep}, a classic residual convolutional network, and \textbf{TinyViT} \cite{wu2022tinyvit}, a
compact vision transformer. Both backbones have approximately 11 million parameters. A classification head, using a single fully connected layer with output dimension equal to the number of classes, is appended to each backbone.

We consider three training regimes: \emph{random initialisation}, where all weights are initialised randomly and updated throughout training from scratch; \emph{full fine-tuning}, where weights are initialised from pre-trained networks and all parameters are updated; \emph{linear probing}, where pre-trained weights are frozen and only the classification head is updated.

\subsection{Data Pre-processing} \label{sec:preprocessing}

All images pass through a unified pre-processing pipeline. When using pre-trained weights, greyscale images are converted to three-channel RGB by replicating the single channel, so as to match the domain of the pre-trained features. This channel replication step is omitted for random-initialisation experiments. All images are resized to $256 \times 256$ pixels for consistency across datasets. Pixels are then cast to \texttt{float32} in the range $[0, 1]$. Augmentations are applied at this stage, after which pixel values are clamped back to $[0, 1]$ to handle any out-of-range values introduced by intensity transformations. Finally, images are normalised using the per-channel mean and standard deviation computed from each specific training dataset; the same dataset-specific statistics are used regardless of whether pre-trained weights are employed, ensuring a consistent normalisation scheme across all experimental conditions.

\section{Experiments and Results} \label{sec:experiments}

\subsection{Datasets} \label{sec:data}

We evaluate on eight datasets from two main domains: \emph{natural image datasets} (CIFAR-10, CIFAR-100 \cite{krizhevsky2009learning}, STL-10 \cite{coates2011analysis}, MNIST \cite{lecun1998gradient}) and \emph{medical image datasets} (BRISC \cite{fateh2026brisc}, BUSI \cite{vallez2025bus}, FETAL-PLANES-DB \cite{burgos2020fetal}, ISIC-2019 \cite{codella2017skin,hernandez2024bcn,tschandl2018ham10000}). Medical imaging benchmarks are particularly relevant because data acquisition and annotation are often expensive, and access to large volumes of labelled data is uncommon, making low-budget performance critical.

\subsection{Experimental Protocol} \label{sec:setup}

For each dataset, we sample 1,000 images using stratified sampling to ensure class balance, following the same controlled low-label protocol across datasets. We allocate 80\% (800) to the training pool, 10\% (100) to validation, and 10\% (100) to the held-out clean test set. For robustness evaluation, every held-out test image is additionally evaluated under every augmentation and available intensity. We refer to the aggregate of the \emph{clean} and systematically augmented views as the \emph{overall} test set; clean performance is reported separately on the 100 original held-out images. Thus, although the clean test contains 100 images by design, the robustness evaluation contains substantially more test views.

The training pool is split into budgets of size $\{10, 20, 50, 100, 200, 500, 800\}$, constructed by cumulatively sampling stratified subsets.  This cumulative design ensures that performance improvements from one budget to the next are attributable solely to the additional labelled samples. Models are retrained independently for each budget level.

Each model is trained for 50 epochs with the Adam optimiser \cite{kingma2015adam} at a learning rate of $10^{-4}$. We set $\lambda = 0.1$ and $\tau = 0.07$ and run five independent trials with different random seeds, reporting mean performance with $\pm 1$ standard deviation intervals. The model checkpoint with the highest macro-averaged F1-score on the validation set is used for test evaluation. ActiveAugment uses a batch size of 16, which is expanded to an effective size of 32 after augmentation; all baselines use a batch size of 32. All experiments are run on either an NVIDIA TITAN RTX (24\,GB VRAM) or a Quadro RTX 6000 (24\,GB VRAM).

We compare against AutoAugment \cite{cubuk2019autoaugment}, RandAugment \cite{cubuk2020randaugment}, and TrivialAugment \cite{muller2021trivialaugment}, all accessed via the \texttt{torchvision} transforms API. For RandAugment, the pipeline length is set to 2, and our intensity levels are mapped to the corresponding magnitude values. TrivialAugment requires no hyperparameter configuration. We also include a \emph{No Augmentation} baseline for reference. All baselines use standard cross-entropy loss with a batch size of 32. % See Supplementary Material for further implementation details.
% For AutoAugment, we select the most appropriate policy from \{\texttt{ImageNet}, \texttt{CIFAR10}, \texttt{SVHN}\} based on domain proximity.

Primarily, we report results using ResNet-18 under the full fine-tuning regime with the BADGE strategy, as it was found to be consistently among the top-performing strategies (see Section \ref{sec:selection_freq}). However, consistent patterns are also observed with TinyViT (under random initialisation and linear probing), as well as in a continual learning setting where increasing budgets are accumulated progressively during training.

\subsection{Results on Natural Image Datasets} \label{sec:natural}

\begin{figure}[t]
    \centering
    \includegraphics[width=1\textwidth]{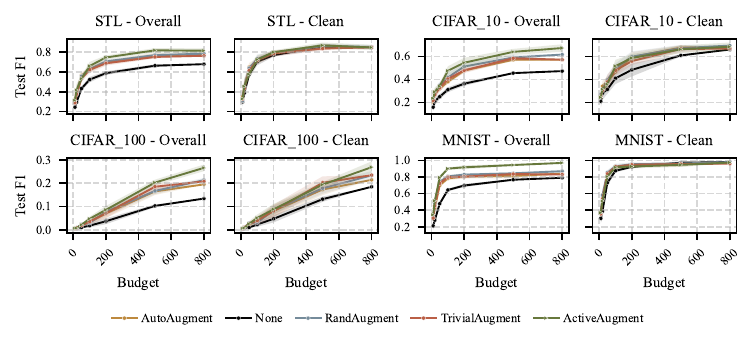}
  \vspace{-0.8cm}
  \caption{Test results on natural image datasets (ResNet-18 backbone, full fine-tuning, BADGE strategy). Shaded bands indicate $\pm 1$ standard deviation. \emph{Overall}: test set including augmented views. \emph{Clean}: original test images only.}
  \label{fig:natural}
\end{figure}

Figure \ref{fig:natural} compares ActiveAugment with the SOTA baselines and the no-augmentation baseline across natural image datasets under full fine-tuning with ResNet-18 and BADGE strategy.

\begin{figure}[t]
    \centering
    \includegraphics[width=1\textwidth]{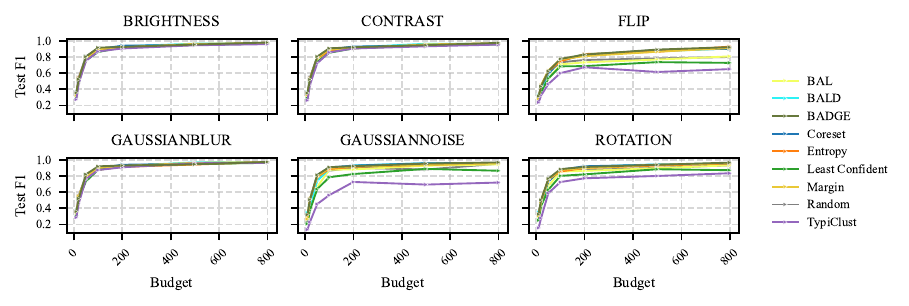}
    \vspace{-0.8cm}
    \caption{Per-augmentation test F1 on MNIST (ResNet-18 backbone, full fine-tuning). BADGE, Margin, Entropy, and Coreset substantially outperform BAL, Least Confident, and TypiClust, with the largest gap observed under the Flip augmentation.}
    \label{fig:TTA_MNIST_resnet}
\end{figure}

On the \emph{overall} test set, ActiveAugment matches or outperforms all baselines. The difference is most pronounced at low budgets: training without augmentations is clearly the weakest strategy, the three SOTA methods perform comparably to one another, and ActiveAugment achieves a small but consistent improvement. The advantage shrinks at larger budgets, as the increased amount of labelled data reduces the relative benefit of adaptive augmentation selection. The pattern is particularly clear on CIFAR-100, where ActiveAugment maintains a visible margin over all SOTA baselines across the full budget range.

On the \emph{clean} test set, the picture is more nuanced. For STL-10 and MNIST, augmentation confers no advantage over the no-augmentation baseline; this is consistent with the nature of these datasets, where the dominant discriminative features are not substantially affected by the augmentations we employ. For CIFAR-10 and CIFAR-100, augmentation clearly improves clean-image performance, and ActiveAugment is competitive with SOTA across all budgets.

These results indicate that the gains from ActiveAugment on the overall test set stem from improved robustness to the specific augmentation types considered, rather than solely from improved clean-image accuracy. It learns to prioritise augmentations under which the model is fragile, thereby improving robustness without sacrificing generalisation to clean data. We additionally test generalisation to held-out transformations by excluding one augmentation type from the training pool and applying it only at test time. ActiveAugment remains competitive with or improves over the random and augmentation baselines, although its gains are smaller than for transformations observed during training.

To illustrate how the selection strategies affect test-time generalisation performance, Figure \ref{fig:TTA_MNIST_resnet} reports the per-augmentation test F1 on MNIST. % Per-augmentation breakdowns for the remaining experiments are provided in Supplementary Material. 
BADGE consistently ranks among the top-performing strategies alongside Margin, Entropy, and Coreset. In contrast, BAL, Least Confident, and
TypiClust consistently underperform, particularly under the Flip and Gaussian Noise augmentations, while BALD and Random also show weaker performance in several settings. The most striking gap appears under the Flip augmentation, where the best strategies reach near-perfect F1 at moderate budgets while BAL, Least Confident, Random and TypiClust lag significantly behind throughout training.

\begin{figure}[t!]
    \centering
    \includegraphics[width=0.9\linewidth,height=0.25\textheight]{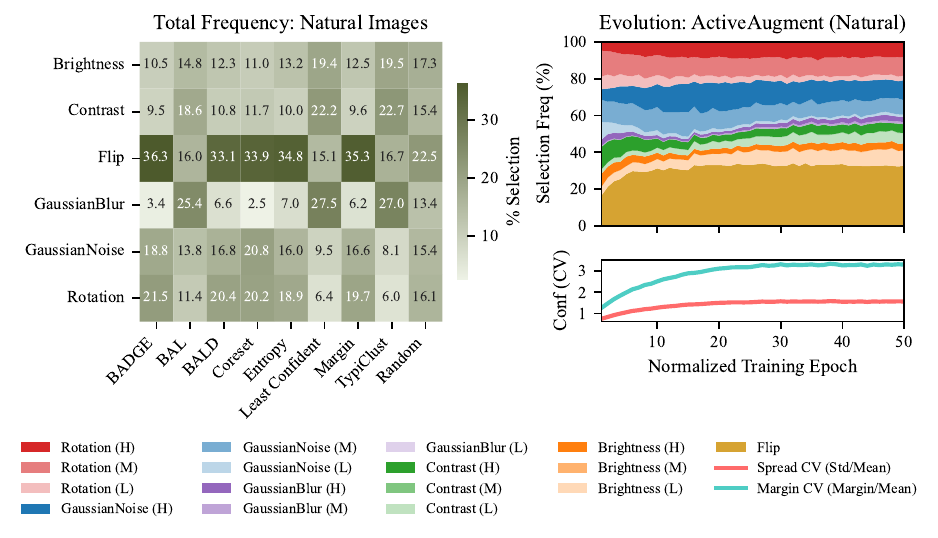}
    \vspace{-0.2cm}
    \caption{\textbf{Left:} Augmentation selection frequency (\%) on natural image datasets (ResNet-18 backbone, full fine-tuning), aggregated over all datasets and budgets. Each column corresponds to one active selection strategy; the rightmost column is the random baseline. \emph{Random} frequencies are non-uniform because the final score is weighted by feature discrepancy. Strategies that favour Gaussian Blur consistently underperform (see Section \ref{sec:natural}). \textbf{Right:} Evolution of augmentation selection frequencies (top) and selection confidence (bottom) over normalised training epochs, aggregated over all natural image datasets and budgets (ResNet-18 backbone, full fine-tuning, BADGE strategy). Confidence is measured as the coefficient of variation of the spread (Std/Mean) and of the margin (Margin/Mean) across augmentation scores. Both confidence metrics rise monotonically, indicating that the selection policy becomes increasingly decisive as the encoder representations stabilise.}
    \label{fig:heatmap_policy_natural}
\end{figure}

\subsection{Augmentation Selection Frequency} \label{sec:selection_freq}

Since ActiveAugment selects a different augmentation for each sample, we can analyse which ones each informativeness strategy prefers. Figure \ref{fig:heatmap_policy_natural} (left) reports selection frequencies aggregated over all natural image datasets and budgets under full fine-tuning with ResNet-18.

\paragraph{Effect of feature discrepancy.}

The random baseline (uniform augmentation sampling) yields non-uniform selection frequencies because the final score is weighted by feature discrepancy. In particular, Flip is over-selected (22.5\% vs.\ the expected 16.7\%), while Gaussian Blur is substantially under-selected (13.4\%). This reveals that horizontal/vertical flipping induces a large representation shift in fine-tuned ResNet-18 features, while Gaussian Blur has comparatively little effect on the embedding, despite being metrically perceptible.

\paragraph{Strategy comparison.}

BADGE, BALD, Coreset, Entropy, and Margin all exhibit a strong preference for Flip, followed by Gaussian Noise and Rotation, and largely ignore Gaussian Blur. In contrast, BAL, Least Confident, and TypiClust preferentially select Gaussian Blur. Crucially, this behavioural difference has a performance consequence: BAL, Least Confident, and TypiClust consistently underperform the other five strategies across datasets, backbones, and training regimes. These results provide strong evidence that Gaussian Blur, despite being perceptibly distinct, does not challenge the model's learned feature representations in a way that promotes generalisation. A good selection strategy should therefore avoid it in favour of transformations that expose genuine model fragilities.

\subsection{Policy Evolution During Training} \label{sec:policy_evolution}

Figure \ref{fig:heatmap_policy_natural} (right) shows how the selection policy of ActiveAugment using BADGE evolves over the course of training, together with a measure of selection confidence. Results are aggregated over all natural image datasets and budgets. % Results for the remaining strategies are provided in Supplementary Material.

We measure confidence using two metrics: the coefficient of variation (CV) of the \emph{spread} (standard deviation over mean score) and the CV of the \emph{margin} (gap between the highest and mean score). At the start of training, confidence is low and the strategy is essentially uncertain about which augmentation to prioritise. As training progresses, confidence rises monotonically, reflecting the model's increasingly stable representations. This emergent learning of a stable selection policy is a direct consequence of the feature discrepancy becoming more informative as the encoder matures.

In terms of augmentation preference, Flip is initially selected less than 20\% of the time but rises to 30--35\% by the end of training, becoming the single most preferred augmentation. At low intensity levels, Gaussian Noise and Rotation are preferred in early epochs but are displaced by Brightness and Contrast augmentations in later stages. Notably, GaussianBlur (L), Brightness (M), and Contrast (M) are virtually never selected across all epochs. Inspection of image corruption metrics confirms that these three configurations have minimal perceptual impact on the image, explaining why the model does not benefit from learning invariance to them. This consistency between the SSIM-based analysis and the selection behaviour provides a meaningful validation of the feature discrepancy component.

\subsection{Results on Medical Image Datasets} \label{sec:medical}

\begin{figure}[t]
  \centering
  \includegraphics[width=1\textwidth]{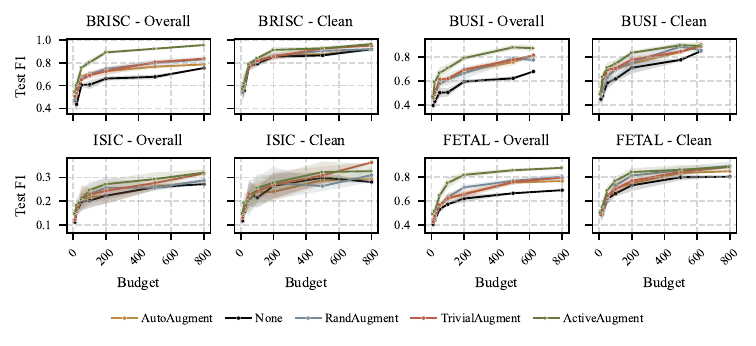}
  % \vspace{-0.25cm}
  \vspace{-0.8cm}
  \caption{Test results on medical image datasets (ResNet-18 backbone, full fine-tuning, BADGE strategy). Shaded bands indicate $\pm 1$ standard deviation. \emph{Overall}: test set including augmented views. \emph{Clean}: original test images only. ActiveAugment is consistently among the top-performing methods on the clean test set across all four datasets and all budgets.}
  \label{fig:medical}
\end{figure}

Figure \ref{fig:medical} compares ActiveAugment with the baselines across the four medical image datasets under full fine-tuning with ResNet-18 and BADGE strategy.

The overall pattern mirrors the natural image results: ActiveAugment achieves a clear advantage on the overall test set, and the gap narrows on the clean test set. On the clean test set, ActiveAugment is consistently among the top-performing methods across all four medical datasets and budget levels, whereas on natural image datasets, the clean-set gap was less pronounced. We attribute this to the out-of-domain nature of these datasets relative to the ImageNet pre-trained models: the SOTA baselines use augmentation policies optimised for natural images (less well-suited to the medical imaging domain), whereas ActiveAugment's data-driven selection adapts automatically to the statistical properties of the target domain. This cross-domain adaptability is a key practical advantage of the proposed framework.

On BRISC, ActiveAugment reaches approximately 0.95 macro-F1 on the clean test set at the highest budget, outperforming all baselines, while on BUSI, the overall and clean curves show a clear separation from AutoAugment and the no-augmentation baseline. On ISIC-2019, which is a challenging dataset due to class imbalance, all methods show wider variance, but ActiveAugment remains competitive even at very low budgets (10--50 samples).

\begin{figure}[t]
    \centering
    \includegraphics[width=1\textwidth]{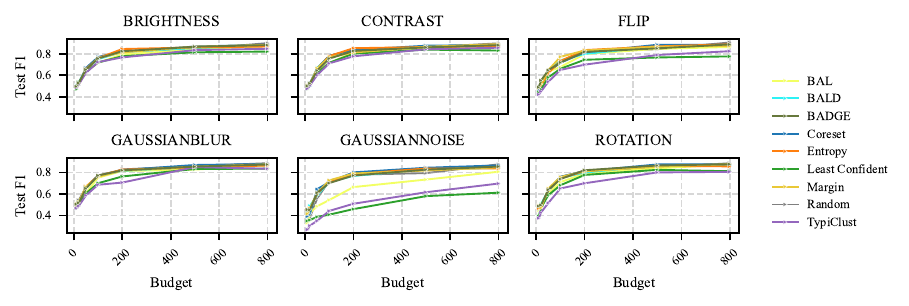}
    \vspace{-0.8cm}
    \caption{Per-augmentation test F1 on FETAL-PLANES (ResNet-18 backbone, full fine-tuning). The performance gap between top and bottom strategies is largest under Gaussian Noise, suggesting that noise robustness is discriminative for medical image recognition.}
    \label{fig:TTA_FETAL_resnet}
\end{figure}

To illustrate how selection strategies affect test-time generalisation performance in the medical domain, Figure \ref{fig:TTA_FETAL_resnet} reports the per-augmentation test F1 on FETAL-PLANES. The ordering of strategies is consistent with the natural image results: BADGE, Margin, Entropy, and Coreset consistently outperform BAL, Least Confident, and TypiClust. The most pronounced gap is observed under Gaussian Noise, reinforcing the finding that noise-based augmentation is a particularly informative perturbation for medical image models.

\begin{figure}[H]
    \centering
    \includegraphics[width=0.9\linewidth,height=0.25\textheight]{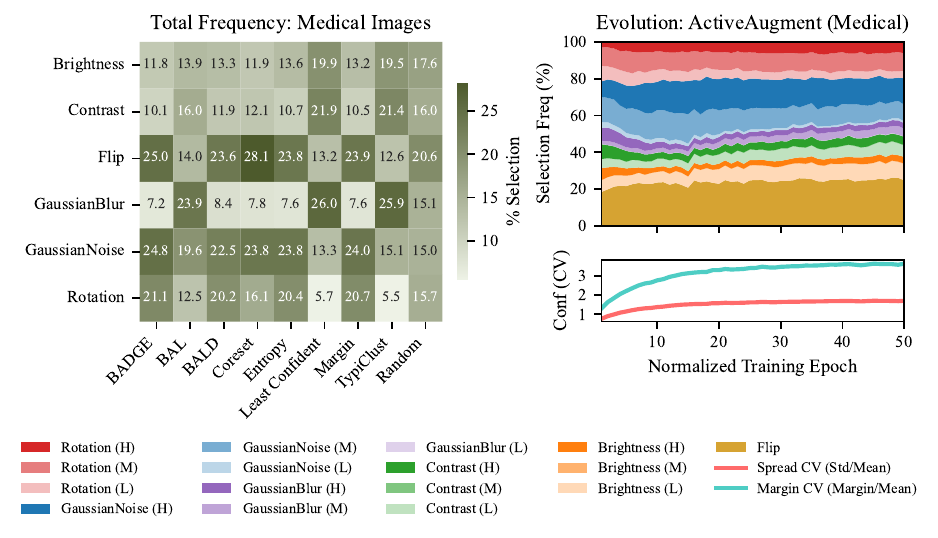}
    \vspace{-0.2cm}
    \caption{\textbf{Left:} Augmentation selection frequency (\%) on medical image datasets (ResNet-18 backbone, full fine-tuning), aggregated over all datasets and budgets. Compared to natural images, Gaussian Noise is selected more frequently (on par with Flip for several strategies). \textbf{Right:} Policy evolution on medical image datasets (ResNet-18 backbone, full fine-tuning, BADGE strategy). Selection confidence rises monotonically as on natural image data, confirming the domain-agnostic behaviour of the selection mechanism.}
    \label{fig:heatmap_policy_medical}
\end{figure}

Selection frequency analysis on medical data (Figure \ref{fig:heatmap_policy_medical} (left)) shows broadly consistent trends with the natural image results. The preference for Flip is somewhat less pronounced, and Gaussian Noise is selected as frequently as Flip by several strategies, suggesting that additive noise is a more discriminative perturbation in the medical imaging domain than in natural imaging. This difference is consistent with the nature of medical images, where structural content is often more orientation-invariant and noise sensitivity is higher.
Policy evolution on medical data (Figure \ref{fig:heatmap_policy_medical} (right)) also shows very similar dynamics to the natural image case, with confidence rising steadily throughout training, confirming that the selection mechanism is domain-agnostic and adapts reliably across image modalities.

\section{Conclusion} \label{sec:conclusion}

We have presented ActiveAugment, a unified framework that reformulates data augmentation selection as an online active learning problem. ActiveAugment scores augmented views per sample using predictive uncertainty and normalised feature discrepancy, and trains the model with a joint classification and supervised contrastive objective that enforces invariance to the selected, model-fragile augmentations. The selection criterion is grounded in domain adaptation theory. By maximising the per-sample product of uncertainty and relative feature shift, ActiveAugment tightens the augmentation-induced generalisation bound at each step.

ActiveAugment consistently outperforms AutoAugment, RandAugment, and TrivialAugment under the controlled augmentation shifts considered, across datasets, backbone architectures, and training regimes, with the most pronounced gains at low labelling budgets. On medical imaging datasets, where ImageNet-pretrained models face a larger domain shift and existing augmentation policies are optimised for natural images, ActiveAugment demonstrates strong cross-domain adaptability by learning the augmentation policy from the evolving model state. Analysis of selection behaviour reveals that the policy evolves meaningfully during training, becoming increasingly decisive as encoder representations stabilise, and that strategy choice directly shapes generalisation: strategies that favour Gaussian Blur underperform those that favour Flip and Gaussian Noise. This constitutes an empirical validation of the feature discrepancy component and provides practical guidance for strategy selection.

\subsection*{Acknowledgements}

This project is supported by the Pioneer Centre for AI, funded by the Danish National Research Foundation (grant number P1).

\newpage
\bibliography{refs}

\end{document}